\documentclass[pmlr]{jmlr} 

\newcommand{\tagdsmode}{nonproceedings}

\makeatletter
\newcommand{\tagdssubmission}{submission}
\newcommand{\tagdsproceedings}{proceedings}

\ifx\tagdsmode\tagdsproceedings

\else\ifx\tagdsmode\tagdssubmission
  \def\ps@jmlrtps{%
    \let\@mkboth\@gobbletwo
    \def\@oddhead{\scriptsize Under Review at the 2nd Conference on Topology, Algebra, and Geometry in Data Science\hfill}%
    \let\@evenhead\@oddhead
    \def\@oddfoot{}%
    \let\@evenfoot\@oddfoot
  }

\else
  \def\ps@jmlrtps{%
    \let\@mkboth\@gobbletwo
    \def\@oddhead{}%
    \let\@evenhead\@oddhead
    \def\@oddfoot{}%
    \let\@evenfoot\@oddfoot
  }
\fi\fi
\makeatother

\usepackage{longtable}

\usepackage{diagbox}
\usepackage{ragged2e}

\usepackage{enumitem}

\usepackage{url}

\newcommand\longvar[1]{\mathchardef\UrlBreakPenalty=100
\mathchardef\UrlBigBreakPenalty=100 #1}

\usepackage{booktabs}
\usepackage[load-configurations=version-1]{siunitx} 

\theorembodyfont{\upshape}
\theoremheaderfont{\scshape}
\theorempostheader{:}
\theoremsep{\newline}

\jmlrvolume{334}
\jmlryear{2026}
\jmlrworkshop{Topology, Algebra, and Geometry in Data Science}

\title[Comparing Chatbot Performance]{Comparing Chatbot Performance Enhanced with Persistent Homology}

\ifx\tagdsmode\tagdssubmission

\else

\author{\Name{Nithisha Raghavaraju} \Email{nithisha2201@gmail.com}\\
\AND
\Name{Barbara Giunti} \Email{bgiunti@albany.edu}\\
\addr SUNY at Albany
\AND
\Name{Bastian Rieck} \Email{bastian.grossenbacher@unifr.ch} \\
\addr{University of Fribourg}
}

\fi

\ifx\tagdsmode\tagdsproceedings
\editor{Editor's name}
\fi

\usepackage{mathtools}
\usepackage{amsfonts}
\usepackage{wasysym}
\usepackage{bbm}
\usepackage{mathbbol} 

\usepackage{hyperref}
\usepackage[noabbrev,nameinlink,capitalise]{cleveref}
\usepackage{thmtools} 
\usepackage{cite}

\usepackage{xcolor}

\usepackage{xargs}

\usepackage{graphicx} 
\usepackage{booktabs} 
\usepackage{multirow}

\usepackage{adjustbox}

\begin{document}

\maketitle

\vspace{-.1cm}
\section{Introduction}\label{sec_intro}

Chatbots have become increasingly prevalent across various domains, offering automated assistance in many areas, especially mental health support. 
They provide an effective alternative to understaffed and overworked healthcare workers, but they need to be properly trained to offer adequate support in the delicate and nuanced realm of mental health. 
The training is done using extremely large datasets, which are sometimes not available in very specific domains. 
Moreover, every person is different, so the model needs to be able to distinguish very subtle differences to understand if the patient is about to cause harm to themselves or others, if they are not at risk now but about to be in the future, etc. 
Thus, it would be ideal to train the chatbot with personal information about the patients, which, of course, cannot be done on shared servers since it would violate patient confidentiality. 
Further, not all facilities have the possibility to train a new chatbot for each need and may be forced to use the same one for very different tasks, thus lowering accuracy. 
Hence, being able to improve the performance of a chatbot, possibly trained locally and on a restricted dataset, without having to increase the dataset itself, would be extremely beneficial. 

Persistent homology (PH) is a tool of Topological Data Analysis (TDA) that is becoming increasingly popular in data analysis for the information it is able to extract from data, which is often orthogonal to what other data analysis methods can retrieve \citep{donut}. 
In short, PH scans the data along all values of a given parameter, and collects a topological signature, called \emph{homology}, at each value. 
It then outputs the evolution of this signature, providing robust insight into the data. 
These outputs can then be vectorized to be made amenable to ML pipelines. 
In this work, we will do exactly that: we will enhance the input datasets using the PH vectorizations computed from the raw datasets themselves. 
Then we will compare, across several metrics, the performance of multiple chatbot models with or without the PH enhancement. 

Our experiments suggest that, while at times the PH enhancement is not particularly beneficial, it sometimes brings remarkable advantages for virtually no cost. 
The most prominent example is the perplexity metric for the decoder-decoder models, where the PH-enhanced chatbots show a consistent and considerable improvement. 
Further work needs to be done to better understand how to improve chatbot performance using PH, since ours is an initial and perhaps naive approach, but our results are encouraging: PH can improve the function of chatbot models at no cost and without needing large datasets.

\section{Models construction and setting}

The overarching goal of this work is to empirically investigate whether topological features derived from Persistent Homology can improve the performance and robustness of conversational AI models. 
We study this question across three distinct architectural paradigms: encoder-only classification models (\emph{encoder-encoder}), sequence-to-sequence generation models (\emph{encoder-decoder}), and autoregressive causal language models (\emph{decoder-decoder}). 
For each paradigm, we compare a baseline model trained solely on text-derived embeddings against an augmented counterpart that additionally receives a fixed-size vector of Topological Data Analysis (TDA) features extracted from the input text. 
To ensure scientific rigor and isolate the impact of the topological signal, the core model capacity, hyperparameter settings, and optimization strategies remain strictly identical between the baseline and TDA-augmented versions.

\subsection{Encoder-encoder}

For the encoder-encoder classification setup, we leverage pre-trained language models to map incoming user utterances to dense contextualized vector embeddings. 
Specifically, we evaluate three robust foundational models. 
First, we use \emph{BERT} (Bidirectional Encoder Representations from Transformers), implemented specifically via the \texttt{\longvar{distilbert-base-uncased}} checkpoint. 
Second, we employ the \emph{Dense Passage Retriever (DPR)}. 
Finally, we evaluate the \emph{Universal Sentence Encoder (USE)}. 
These pre-trained models act as frozen feature extractors, and their textual embeddings serve as the primary input for a downstream shallow Multi-Layer Perceptron (MLP) classification block.

Formally, let $\mathbf{h} \in \mathbb{R}^{d}$ denote the pre-computed text embedding (e.g., the \texttt{[CLS]} token embedding, where $d = 768$ for BERT and DPR, and $d = 512$ for USE). The text embedding is passed through a fully connected projection block:
\[
  \mathbf{p} = \mathrm{Dropout}_{0.3}\left(\mathrm{ReLU}\left(\mathrm{LN}(\mathbf{W}_p \mathbf{h} + \mathbf{b}_p)\right)\right) \in \mathbb{R}^{128},
\]
where $\mathrm{LN}$ denotes Layer Normalization, and $\mathbf{W}_p$ is a learnable projection matrix. 

When topological features are incorporated, the PH pipeline yields a feature vector $\mathbf{t} \in \mathbb{R}^{d_{\mathrm{PH}}}$ (with $d_{\mathrm{PH}} = 100$, obtained via Principal Component Analysis of persistence landscapes built upon FastText word embeddings). This vector is similarly projected:
\[
  \mathbf{q} = \mathrm{ReLU}(\mathrm{LN}(\mathbf{W}_q \mathbf{t})) \in \mathbb{R}^{128}.
\]
The text and PH projections are then fused using a learned gating mechanism. This mechanism computes a soft convex combination of the representations:
\[
  \boldsymbol{\alpha} = \sigma\left(\mathbf{W}_g [\mathbf{p};\mathbf{q}]\right), \qquad \mathbf{f} = \boldsymbol{\alpha} \odot \mathbf{p} + (\mathbf{1} - \boldsymbol{\alpha}) \odot \mathbf{q},
\]
where $[\cdot;\cdot]$ denotes vector concatenation, $\sigma$ is the element-wise sigmoid function, and $\odot$ is the Hadamard product. The fused representation connects to the final classifier block. Crucially, the concatenated vector $[\mathbf{p};\mathbf{f}] \in \mathbb{R}^{256}$ is fed into the final classifier layer in both conditions; for the purely textual baseline, we set $\mathbf{f} = \mathbf{0}$, thus maintaining identical mathematical capacity within the classifier head.

Depending on the targeted dataset, the model acts as a multi-label or single-label classifier. For single-label tasks, the network minimizes the cross-entropy loss, and validation is assessed via accuracy. For multi-label scenarios, binary cross-entropy is utilized, and the model's performance is evaluated using comprehensive metrics: F1-score (Micro and Macro), Precision, Recall, and AUC. All encoder-encoder configurations are trained using the AdamW optimizer (learning rate $10^{-3}$, batch size 16) with a step-wise learning rate scheduler and an early stopping mechanism (patience of 3 epochs).

\subsection{Encoder-decoder}

The encoder-decoder architecture forms the classical backbone for open-ended chatbot response generation, mapping a user's input sequence to an appropriate conversational reply. We adapt four distinct sequence transduction models of varying inductive bias: Recurrent Neural Networks (LSTM and GRU) and Transformer-based models (a custom T4 Transformer and pre-trained BART).

\paragraph{Recurrent architectures (LSTM and GRU).} 
For the \emph{Long Short-Term Memory (LSTM)} model, tokenized input sequences are represented as word embeddings $\mathbf{e}_t \in \mathbb{R}^{E}$ ($E=256$). 
When PH is enabled, the 10-dimensional PH vector $\mathbf{t} \in \mathbb{R}^{10}$ is broadcast across the sequence length and channel-wise concatenated to the embeddings: $\tilde{\mathbf{e}}_t = [\mathbf{e}_t;\, \mathbf{t}] \in \mathbb{R}^{E + 10}$. This represents the sole architectural variation. The bidirectional LSTM encoder (hidden dimension $H=512$) condenses the sequence into a final hidden state that initializes the unidirectional LSTM decoder.
The \emph{Gated Recurrent Unit (GRU)} model follows an identical concatenation strategy at the embedding layer. However, its decoder employs a Bahdanau-style additive attention over the encoder's intermediate sequence states. During inference, both recurrent models utilize teacher-forcing during training and greedily decode the response. They are trained across 100 epochs using Adam ($10^{-3}$), equipped with gradient clipping and early stopping monitored on validation BLEU scores.

\paragraph{Transformer architectures (T4 and BART).}
The \emph{T4 Transformer} is a custom-built, randomly initialized sequence-to-sequence model featuring 4 encoder and decoder layers, 8 attention heads, and a fixed model dimension of $d_{\mathrm{model}} = 256$. 
To enforce strict ablation parity without altering the sub-layer dimensions, the PH features $\mathbf{t} \in \mathbb{R}^{8}$ are injected via a tailored linear projection $\mathbf{W}_{\mathrm{PH}} \in \mathbb{R}^{d_{\text{model}} \times \text{dim}(\mathbf{t})}$. 
This mapped representation is then broadcast and added element-wise to the embedded source tokens prior to sinusoidal positional encoding computation.
Conversely, \emph{BART} (\texttt{facebook/bart-base}) is a heavily parameterized ($\approx 140$M parameters) pre-trained denoising autoencoder. 
We introduce PH features by linearly projecting the 10-dimensional topological vector to BART's native hidden dimension ($d_{\text{model}} = 768$) and additively injecting it into every position of the encoder's terminal hidden state before cross-attention computation in the decoder. 
This lightweight intervention adds a negligible $7{,}680$ parameters, effectively preserving parameter equivalence. BART fine-tuning relies on AdamW (learning rate $5\times10^{-5}$) and uses BERTScore F1 as the primary validation metric under an identical early stopping policy.

\subsection{Decoder-decoder}

The decoder-decoder regime models the conversational generation objective as a purely autoregressive language modeling task, eliminating the distinct geometric separation of the encoder. The generative models are expected to implicitly condition their response generation solely on the left-to-right causal attention prefix. 

We assess four modern foundational language models of varying scale and architecture: \texttt{distilgpt2} ($\approx 82$M parameters), \texttt{gpt2-medium} ($\approx 345$M parameters), \longvar{TinyLlama-1.1B-Chat-v1.0} ($\approx 1.1$B parameters), and \texttt{Qwen1.5-0.5B-Chat} ($\approx 0.5$B parameters). 
To manage the computational footprint of the larger models (i.e., TinyLlama and Qwen), we employ parameter-efficient fine-tuning via Low-Rank Adaptation (LoRA). 
The LoRA adapters target the query (\texttt{q\_proj}) and value (\texttt{v\_proj}) self-attention matrices with a minimal rank constraint ($r=8$) and alpha scaling factor ($\alpha=16$), heavily reducing memory overhead while preserving convergence quality. 
All models interpret conversational context structured by specific chat templates (e.g., encapsulating the prefix within \texttt{<|user|>} and \texttt{<|assistant|>} tokens).

Because these models lack a dedicated encoder hidden dimension for arithmetic feature fusion, topological descriptors must be seamlessly serialized. 
During the data preprocessing workflow, the 10-dimensional PCA-reduced topological landscape array is converted into a pseudo-lexical string payload. Each scalar coefficient $v_i$ is mapped to a discrete, format-consistent string identifier (e.g., \texttt{<tda0:0.123>} \texttt{<tda1:-0.456>}). 
This structured topological sequence is appended directly to the user's natural language input within the prompt bounding. 
Thus, the autoregressive transformer discovers the cross-modal correlations between the topological prompt tokens and the lexical content via standard causal self-attention mechanisms. Evaluation across the decoder landscape leverages generation metrics including 
ROUGE-L.

\section{Experiments}

\subsection{Persistent homology setup}\label{ssec_ph_setup}

Persistent homology (PH) is a widely used tool in Topological Data Analysis (TDA), with over 350 applications \citep{donut}. 
It extracts topological features at different scales along one varying parameter, outputting their lifespans (or \emph{persistence}) with respect to the parameter's values. 
The output is thus a collection of intervals, equivalently seen as points in the plane, with multiplicity.
Such an output is not amenable to being passed as input for training, and thus we vectorize it using \emph{persistence landscapes} \citep{Bubenik2015}. 
We refer to standard texts in general persistence theory \citep{Dey_Wang_2022} and in persistence theory for machine learning \citep{Hensel21, Zia24a} for more details. 

\begin{table}[ht]
\centering
\renewcommand{\arraystretch}{1.3}
\adjustbox{scale=.9}{
\begin{tabular}{
>{\RaggedRight\arraybackslash}p{3.7cm}|
>{\RaggedRight\arraybackslash}p{4.5cm}|
>{\RaggedRight\arraybackslash}p{3.7cm}|
>{\RaggedRight\arraybackslash}p{3.37cm}}

\diagbox[width=10.7em,innerleftsep=0pt, innerrightsep=5pt, height=2.5em]{\adjustbox{scale=.9}{Parameter}}{\adjustbox{scale=.9}{Model}} &
\textbf{Encoder-Encoder} &
\textbf{Encoder-Decoder} &
\textbf{Decoder-Decoder} \\
\hline

FastText dimension $d$ &
100 &
50 &
100 \\
\hline

Filtration threshold $r_{\max}$ &
Dynamic: 90th percentile of pairwise distances, fallback 0.1 &
Dynamic: 90th percentile of pairwise distances, fallback 0.1 &
Fixed: $r_{\max}=10$ \\
\hline

Landscape resolution $R$ &
200 &
100 &
50 \\
\hline

PCA Components &
100 (300 for USE models) &
10 &
250 \\
\end{tabular}
}
\caption{Comparison of parameter settings across model architectures.}
\label{tab:model_params}
\end{table}

Word embeddings were obtained using the Gensim library \citep{rehurek_lrec} to train a FastText model on the tokenized training splits.
Text tokenization was performed using the NLTK library \citep{nltk}, capping at a maximum of 100 tokens for computational tractability. 
We trained a FastText model on the tokenized training split of the dataset. 
For each sentence, we retrieved the FastText word vectors for all tokens to construct a point cloud in $\mathbb{R}^d$ of size $(N,d)$, where $N\leq 100$ is the token count and $d$ is the embedding dimension.
Edge cases were handled as follows: out-of-vocabulary or missing tokens defaulted to zero vectors. 
For sentences containing only a single word, a zero vector of size d was appended to ensure a minimum of 2 points, enabling distance computations.

To obtain the PH vectors, we used the GUDHI library \citep{gudhi} as follows.
From the embeddings, we calculated all pairwise Euclidean distances, and constructed the Vietoris--Rips filtration from them, with a maximum filtration edge length to avoid extremely sparse components. 
Then we computed the raw persistence diagrams, which were vectorized with the following pipeline: 
\begin{itemize}
    \item Diagram Selector: Filtered out infinite birth-death points, keeping only finite topological features.
    \item Diagram Scaler: Scaled and normalized the birth-death coordinates.
    \item Clamping: Applied clamping bounds (maximum threshold of 0.9) to the death coordinates to prevent extreme outliers from skewing the representations.
    \item Landscape: Constructed persistent landscapes (representing functions of birth-death curves) using a fixed resolution parameter $R$.
\end{itemize}

The parameters used are collected in \cref{tab:model_params}.
If a text had no persistence points, i.e., an empty diagram, it was mapped directly to a zero vector of the corresponding target length.

The raw persistent landscapes output by GUDHI were flattened, and sizing discrepancies arising from empty diagrams or different numbers of landscape functions were aligned via zero-padding or truncation to ensure uniform vector size.
To compress the topological features and avoid the curse of dimensionality during neural network integration, we applied Principal Component Analysis (PCA) to reduce the flattened landscape vectors to a fixed number of principal components.
These steps were performed using scikit-learn \citep{scikit-learn}. 
To maintain strict experimental rigor, the PCA model and subsequent StandardScaler were fitted only on the training split's PH vectors, and then applied to transform the validation and test splits.
\medskip 

While all models compute the PH features using the same steps presented thus far, they differ in how they ``consume'' those features. 
For the Encoder-Encoder models and the Encoder-Decoder models, inside the PyTorch model's forward pass, the PH vectors are mathematically concatenated to the hidden layer embeddings. 
The model handles them purely as numeric features.

For the Decoder-Decoder family, PH features were integrated using a text-prompt injection method. 
In detail, the mean FastText word embeddings of the sentence (100 dimensions) and the PCA-reduced PH landscapes (250 dimensions) were individually scaled.    The scaled vectors were concatenated into a single 350-dimensional vector. 
The first 10 dimensions of this combined representation were extracted.
This compact vector was formatted into custom text tokens 
and appended to the user text input in the prompt template, allowing the Decoder-Decoder models to process the topological features directly as part of their natural language context during fine-tuning.

\subsection{Datasets}

A diverse mental health NLP corpus was constructed from eight publicly available sources, partitioned so that each architecture is evaluated on exactly four datasets:
\begin{itemize}
    \item \textbf{Encoder-Encoder (Classification Models):} Four classification datasets mapping text to categories, emotions, or clinical indicators.
    \item \textbf{Encoder-Decoder (Generative Models):} Four dialogue datasets mapping inputs to sequence-to-sequence response text.
    \item \textbf{Decoder-Decoder (Causal Models):} The same four dialogue datasets, formatted as a single-sequence prompt.
\end{itemize}

\textbf{Conversational/Dialogue Datasets} (used by Encoder-Decoder and Decoder-Decoder models): CounselChat QA \citep{bertagnolli2020counsel}, MentalChat16K \citep{shen2023mental,chia2023mentalchat}, Mental Health Chatbot Dataset \citep{brahma2023mental}, and NLP-4-Mental-Health \citep{fedestack2022nlp}.

\textbf{Classification/Labelled Datasets} (used by Encoder-Encoder models): CounselChat Topics \citep{bertagnolli2020counsel}, DepressionEmo \citep{rahman2024depression}, Reddit Mental Health Dataset \citep{entenam2021reddit}, and NLP-4-Mental-Health Combined \citep{fedestack2022nlp}.

\textbf{Preprocessing:} Encoder-Encoder uses datasets with integer or multi-hot encoding. 
Encoder-Decoder retains only input $X$ and response $Y$. 
Decoder-Decoder merges input and output into a single sequence with speaker delimiters (for example, \texttt{<|user|>:} \texttt{\{context\}\textbackslash n} \texttt{<|assistant|>:\{response\}}) and applies a prompt mask so loss is computed on assistant tokens only.

\subsection{Evaluation Metrics}\label{ssec_metrics}

The choice of evaluation metrics is tailored to the task type of each dataset:
\begin{itemize}
    \item \textbf{Classification Metrics (F1, Accuracy, AUC):} For the labeled classification datasets (DepressionEmo \citep{rahman2024depression}, Reddit Mental Health Dataset \citep{entenam2021reddit}), we use standard classification metrics to measure the model's ability to categorize text into discrete emotional states or topics.
    \item \textbf{Generation Metrics (BLEU, ROUGE-L, BERTScore):} For the conversational datasets (CounselChat QA \citep{bertagnolli2020counsel}, MentalChat16K \citep{shen2023mental,chia2023mentalchat}), response generation is framed as a sequence-to-sequence mapping (similar to machine translation). Thus, we use BLEU and ROUGE-L to measure lexical overlap between the generated response and the reference therapist reply. 
    To capture semantic similarity beyond exact word matches, we also use BERTScore.
    \item \textbf{Language Modeling Metric (Perplexity):} For the decoder-decoder models trained on the dialogue datasets, perplexity (PPL) is used to evaluate how well the autoregressive model has learned the language distribution of the clinical dialogue.
\end{itemize}

\subsection{Results and discussion}

The tests were run on a workstation with a 13th Gen Intel(R) Core(TM) i7-13700 CPU and 32 GB of RAM, running Windows 11 Enterprise, with Python version 3.13.13. 
The full code for both training and experiments, including databases, is available at \citet{github_repo_encenc,github_repo_encdec,github_repo_decdec}. 
All results are the average over five repetitions with different (fixed) random seeds (42, 43, 44, 45, 46).

\subsubsection{Encoder-encoder}

When using encoder-encoder architectures, we observe that the addition of topological features is largely \emph{neutral}, in the sense that we observe no substantial improvements in performance metrics. This pattern is consistent over different model; cf.\ \Cref{tbl:enc-enc_bert,tbl:enc-enc_dpr,tbl:enc-enc_use}. 

We reported the output of the BERT model as exemplary here, and the outputs of the DPR and USE models in the appendix.

\begin{table}[h!]
\begin{adjustbox}{width=\columnwidth,center}
\begin{tabular}[t]{lrrrrrrrr}
\toprule
\multirow{2}{*}{BERT} & \multicolumn{2}{c}{CounselChat} & \multicolumn{2}{c}{Depression}  & \multicolumn{2}{c}{MentalHealth} & \multicolumn{2}{c}{Suicide} \\
\cmidrule(l{3pt}r{3pt}){2-3} \cmidrule(l{3pt}r{3pt}){4-5} \cmidrule(l{3pt}r{3pt}){6-7} \cmidrule(l{3pt}r{3pt}){8-9}
 & Raw & PH & Raw & PH & Raw & PH & Raw & PH \\
\midrule 
Loss  & 1.0425 & 1.0533 & 0.60054 & 0.6078 & 0.9078 & 0.9012 & 0.2585 & 0.2338 \\ 
F1-micro  & 0.3294 & 0.3347 & 0.6057 & 0.6027 & - & - & - & - \\ 
F1-macro  & 0.1817 & 0.1752 & 0.5759 & 0.5731 & - & - & - & - \\  
Precision  & 0.2560 & 0.2606 & 0.6747 & 0.6714 & - & - & - & - \\ 
Recall  & 0.4602 & 0.4678 & 0.5494 & 0.5467 & - & - & - & - \\ 
AUC  & 0.8430 & 0.8434 & 0.7973 & 0.7958 & - & - & - & - \\ 
Acc  & - & - & - & - & 0.6620 & 0.6743 & 0.9316 & 0.9305 \\  
\bottomrule
\end{tabular}
\end{adjustbox}
\caption{Encoder-encoder BERT model, all datasets and metrics.}
\label{tbl:enc-enc_bert}
\end{table}

\subsubsection{Encoder-decoder}

For the encoder-decoder architectures, we observe a more nuanced pattern. 
First, as \Cref{tbl:enc-dec_loss} shows, the overall loss is only marginally affected by the inclusion of PH features, thus indicating that the inclusion of such features is, in the worst case, still \emph{neutral} for a model and does not prevent it from learning. 
With respect to the BLEU metric, a PH-enhanced model exhibits \emph{substantial improvements} in many datasets. 
As \Cref{tbl:enc-dec_bleu} shows, the BART model does not appear to be able to gainfully use PH features, unlike all other models.

\begin{table}[h!]
\begin{adjustbox}{width=\columnwidth,center}
\begin{tabular}[t]{lrrrrrrrr}
\toprule
\multirow{2}{*}{Loss} & \multicolumn{2}{c}{MentalChat} & \multicolumn{2}{c}{CounselChat}  & \multicolumn{2}{c}{MentalChat} & \multicolumn{2}{c}{NLPMentalHealth} \\
\cmidrule(l{3pt}r{3pt}){2-3} \cmidrule(l{3pt}r{3pt}){4-5} \cmidrule(l{3pt}r{3pt}){6-7} \cmidrule(l{3pt}r{3pt}){8-9}
 & Raw & PH & Raw & PH & Raw & PH & Raw & PH \\
\midrule
BART   & 3.7093 & 6.7423 & 3.3173 & 4.7234 & 5.0160 & 7.6780 & 6.0062 & 7.6734 \\
LSTM   & 13.690 & 14.149 & 8.243 & 7.746 & 7.625 & 7.479 & 13.239 & 12.479 \\
GRU   & 738.62 & 748.86 & 565.62 & 580.06 & 657.09 & 658.81 & 679.60 & 695.23 \\
T4    & 4.9495 & 5.0187 & 6.0750 & 6.0866 & 7.1191 & 7.0699 & 7.6231 & 7.6529 \\
\bottomrule
\end{tabular}
\end{adjustbox}
\caption{Loss metric for all encoder-decoder models and all datasets.}
\label{tbl:enc-dec_loss}
\end{table}

\begin{table}[h!]
\begin{adjustbox}{width=\columnwidth,center}
\begin{tabular}[t]{lrrrrrrrr}
\toprule
\multirow{2}{*}{BLEU} & \multicolumn{2}{c}{MentalChat} & \multicolumn{2}{c}{CounselChat}  & \multicolumn{2}{c}{MentalChat} & \multicolumn{2}{c}{NLPMentalHealth} \\
\cmidrule(l{3pt}r{3pt}){2-3} \cmidrule(l{3pt}r{3pt}){4-5} \cmidrule(l{3pt}r{3pt}){6-7} \cmidrule(l{3pt}r{3pt}){8-9}
 & Raw & PH & Raw & PH & Raw & PH & Raw & PH \\
\midrule
BART   & 0.4526 & 0.0411 & 0.3813 & 0.3271 & 0.0697 & 0.0317 & 0.0901 & 0.0311 \\
LSTM   & 0.4021 & 0.4240 & 0.1149 & 0.0336 & 0.0225 & 0.0819 & 0.0624 & 0.0771 \\
GRU   & 0.2115 & 0.2521 & 0.0989 & 0.0565 & 0.0620 & 0.0316 & 0.0226 & 0.0458 \\
T4   & 0.4743 & 0.2466 & 0.1850 & 0.1822 & 0.0272 & 0.0624 & 0.0023 & 0.0131 \\
\bottomrule
\end{tabular}
\end{adjustbox}
\caption{BLEU metric for all encoder-decoder models and all datasets.}
\label{tbl:enc-dec_bleu}
\end{table}

\subsubsection{Decoder-decoder}

For the decoder-decoder models, a surprising finding is that the presence of PH features \emph{substantially} improves the perplexity metric; cf.\ \Cref{tbl:dec-dec_ppl}. 
That is, the perplexity score of PH-enhanced models is substantially lower than for the baseline model. 
This demonstrates that, despite the conceptual simplicity of our inclusion procedure, PH features are capable of improving the learning process of large language models. 
A possible explanation for this behavior is that PH can capture complementary information about the geometry and topology of the respective latent space. 
Making a model ``aware'' of such features thus appears to be a promising avenue for future research.

\begin{table}[h!]
\begin{adjustbox}{width=\columnwidth,center}
\begin{tabular}[t]{lrrrrrrrr}
\toprule
\multirow{2}{*}{PPL} & \multicolumn{2}{c}{CounselChat} & \multicolumn{2}{c}{MH-Chatbot}  & \multicolumn{2}{c}{MentalChat} & \multicolumn{2}{c}{NLP-MH} \\
\cmidrule(l{3pt}r{3pt}){2-3} \cmidrule(l{3pt}r{3pt}){4-5} \cmidrule(l{3pt}r{3pt}){6-7} \cmidrule(l{3pt}r{3pt}){8-9}
 & Raw & PH & Raw & PH & Raw & PH & Raw & PH \\
\midrule
distilgpt2  & 22.83 & 9.18 & 76.33 & 14.34 & 89.71 & 20.12 & 42.76 & 9.69 \\
gpt2-medium  & 14.40 & 6.78 & 45.68 & 10.56 & 56.29 & 14.72 & 33.90 & 8.39 \\
TinyLlama  & 8.39 & 3.97 & 14.60 & 4.86 & 21.78 & 6.79 & 15.27 & 4.78 \\
QWEN   & 13.92 & 5.29 & 39.38 & 7.77 & 62.23 & 11.52 & 34.16 & 6.95 \\
\bottomrule
\end{tabular}
\end{adjustbox}
\caption{PPL metric all decoder-decoder models and all datasets.}
\label{tbl:dec-dec_ppl}
\end{table}

According to some of the other metrics, the addition of PH seems not to bring exceptional improvement. 
In terms of the ROUGE metric, there is only minor improvements with PH features, except in the GWEN model, which seems to benefit across all but one dataset.
This may of course mean that PH is not extremely useful for improving these metrics.
However, since PPL, which is arguably the metric to watch for unsupervised text generation, improves substantially with PH features, this also illustrates the need for a better understanding of evaluation techniques. 
For example, the ROUGE score requires reference text, meaning that the comparison could be biased: A topology-aware model could generate text that is not captured by the reference, resulting in penalties.
%

\newpage
\bibliography{chatbots_article}

@article{Bubenik2015,
  title={Statistical topological data analysis using persistence landscapes},
  author={Peter Bubenik},
  journal={Journal of Machine Learning Research},
  volume={16},
  pages={77--102},
  year={2015},
  url={https://doi.org/10.48550/arXiv.1207.6437}
}

@inproceedings{rehurek_lrec,
title = {{Software Framework for Topic Modelling with Large Corpora}},
author = {Radim {\v R}eh{\r u}{\v r}ek and Petr Sojka},
booktitle = {{Proceedings of the LREC 2010 Workshop on New Challenges for NLP Frameworks}},
pages = {45--50},
year = 2010,
publisher = {ELRA},
address = {Valletta, Malta},
note={\url{http://is.muni.cz/publication/884893/en}},
language={English}
}

@misc{nltk,
  author = {\url{www.nltk.org}},
  title = {Natural Language Toolkit},
  url = {http://www.nltk.org/},
  year = 2012
}

@article{scikit-learn,
  title={Scikit-learn: Machine Learning in {P}ython},
  author={Pedregosa, F. and Varoquaux, G. and Gramfort, A. and Michel, V. and Thirion, B. and Grisel, O. and Blondel, M. and Prettenhofer, P. and Weiss, R. and Dubourg, V. and Vanderplas, J. and Passos, A. and Cournapeau, D. and Brucher, M. and Perrot, M. and Duchesnay, E.},
  journal={Journal of Machine Learning Research},
  volume={12},
  pages={2825--2830},
  year={2011}
}

@book{gudhi, 
title   = "{GUDHI} User and Reference Manual", 
author  = "{The GUDHI Project}", 
publisher     = "{GUDHI Editorial Board}", 
type = {Software},
year         = 2015, 
url =    "http://gudhi.gforge.inria.fr/doc/latest/"
}

@book{Dey_Wang_2022, 
    place={Cambridge}, 
    title={Computational Topology for Data Analysis}, 
    publisher={Cambridge University Press}, 
    author={Dey, Tamal Krishna and Wang, Yusu}, 
    year={2022}
}

@misc{donut,
	author = {Giunti, Barbara and Lazovskis, J{\=a}nis and Rieck, Bastian},
	title  = {
	{DONUT}: {D}atabase of {O}riginal \& {N}on-{T}heoretical {U}ses of {T}opology},
	note   = {\url{https://donut.topology.rocks}},
	year   = {2022},
	key    = {DONUT},
}

@misc{bertagnolli2020counsel,
    author = {Bertagnolli, Nick},
    title = {Counsel Chat: Bootstrapping High-Quality Therapy Data},
    year = {2020},
    publisher = {GitHub},
    url = {https://github.com/nbertagnolli/counsel-chat}
}

@misc{shen2023mental,
    author = {Shen, Yihe and Yang, Zhengyuan and Chen, Xuhai and Wu, Jianfei},
    title = {MentalChat16K: A Multi-Turn Mental Health Dialogue Dataset},
    year = {2023},
    publisher = {HuggingFace},
    url = {https://huggingface.co/datasets/ShenLab/MentalChat16K}
}

@misc{github_repo_encenc,
	author = {Nithisha Raghavaraju},
	title = {{GitHub repository with code for the Encoder-Encoder models and experiments of ``Comparing Chatbot Performance Enhanced with Persistent
Homology''}},
	year = {2026},
	type = {Software},
	publisher = {Github},
	url = {https://github.com/raghavarajunithisha-lab/Encoder-Encoder},
}

@misc{github_repo_encdec,
	author = {Nithisha Raghavaraju},
	title = {{GitHub repository with code for the Encoder-Decoder models and experiments of ``Comparing Chatbot Performance Enhanced with Persistent
Homology''}},
	year = {2026},
	type = {Software},
	publisher = {Github},
	url = {https://github.com/raghavarajunithisha-lab/Encoder-Decoder},
}

@misc{github_repo_decdec,
	author = {Nithisha Raghavaraju},
	title = {{GitHub repository with code for the Decoder-Decoder models and experiments of ``Comparing Chatbot Performance Enhanced with Persistent
Homology''}},
	year = {2026},
	type = {Software},
	publisher = {Github},
	url = {https://github.com/raghavarajunithisha-lab/decoder-decoder},
}

@misc{chia2023mentalchat,
    author = {Chia, Patricia},
    title = {MentalChat16K Main},
    year = {2023},
    publisher = {GitHub},
    url = {https://github.com/ChiaPatricia/MentalChat16K_Main}
}

@misc{brahma2023mental,
    author = {Brahma, Helio},
    title = {Mental Health Chatbot Dataset},
    year = {2023},
    publisher = {HuggingFace},
    url = {https://huggingface.co/datasets/heliosbrahma/mental_health_chatbot_dataset}
}

@misc{fedestack2022nlp,
    author = {Stack, Federico},
    title = {NLP-4-Mental-Health: Datasets and Models for Psychiatric NLP},
    year = {2022},
    publisher = {GitHub},
    url = {https://github.com/Fede-stack/NLP-4-Mental-Health}
}

@misc{rahman2024depression,
    author = {Rahman, Abu Bakar Siddiqur and Chakraborty, Sourav and Suhara, Yoshihiko and Jeffery, Adam and Abdelwahab, Ahmed},
    title = {DepressionEmo: A Novel Dataset for Multilabel Classification of Depression Emotions},
    year = {2024},
    publisher = {GitHub},
    url = {https://github.com/abuBakarSiddiqurRahman/DepressionEmo}
}

@misc{entenam2021reddit,
    author = {Entenam, Mahdi},
    title = {Reddit Mental Health Dataset},
    year = {2021},
    publisher = {Kaggle},
    url = {https://www.kaggle.com/datasets/entenam/reddit-mental-health-dataset/data}
}

@article{Hensel21,
  title        = {A Survey of Topological Machine Learning Methods},
  author       = {Hensel, Felix and Moor, Michael and Rieck, Bastian},
  author+an    = {3=highlight},
  year         = 2021,
  journal      = {Frontiers in Artificial Intelligence},
  volume       = 4,
  doi          = {10.3389/frai.2021.681108},
  issn         = {2624-8212},
}

@article{Zia24a,
  title         = {Topological deep learning: a review of an emerging paradigm},
  author        = {Zia, Ali and Khamis, Abdelwahed and Nichols, James and Tayab, Usman Bashir and Hayder, Zeeshan and Rolland, Vivien and Stone, Eric and Petersson, Lars},
  year          = 2024,
  day           = 29,
  journal       = {Artificial Intelligence Review},
  volume        = 57,
  number        = 4,
  pages         = 77,
  doi           = {10.1007/s10462-024-10710-9},
}

\appendix 

\section{Full experiements results}

\begin{table}[h!]
\begin{adjustbox}{width=\columnwidth,center}
\begin{tabular}[t]{lrrrrrrrr}
\toprule
\multirow{2}{*}{DPR} & \multicolumn{2}{c}{CounselChat} & \multicolumn{2}{c}{Depression}  & \multicolumn{2}{c}{MentalHealth} & \multicolumn{2}{c}{Suicide} \\
\cmidrule(l{3pt}r{3pt}){2-3} \cmidrule(l{3pt}r{3pt}){4-5} \cmidrule(l{3pt}r{3pt}){6-7} \cmidrule(l{3pt}r{3pt}){8-9}
 & Raw & PH & Raw & PH & Raw & PH & Raw & PH \\
\midrule 
Loss   & 0.9849 & 1.0173 & 0.6015 & 0.6108 & 0.9445 & 0.9543 & 0.3065 & 0.3133 \\ 
F1-micro  & 0.3928 & 0.3689 & 0.6105 & 0.6085 & - & - & - & - \\ 
F1-macro   & 0.2311 & 0.2037 & 0.5830 & 0.5813 & - & - & - & - \\  
Precision   & 0.3058 & 0.2872 & 0.6801 & 0.6779 & - & - & - & - \\ 
Recall   & 0.5490 & 0.5155 & 0.5538 & 0.5521 & - & - & - & - \\ 
AUC  & 0.8683 & 0.8606 & 0.8021 & 0.8022 & - & - & - & - \\ 
Acc  & - & - & - & - & 0.6631 & 0.6614 & 0.9144 & 0.9213 \\  
\bottomrule
\end{tabular}
\end{adjustbox}
\caption{Encoder-encoder DPR model, all datasets and metrics.}
\label{tbl:enc-enc_dpr}
\end{table}

\begin{table}[h!]
\begin{adjustbox}{width=\columnwidth,center}
\begin{tabular}[t]{lrrrrrrrr}
\toprule
\multirow{2}{*}{USE} & \multicolumn{2}{c}{CounselChat} & \multicolumn{2}{c}{Depression}  & \multicolumn{2}{c}{MentalHealth} & \multicolumn{2}{c}{Suicide} \\
\cmidrule(l{3pt}r{3pt}){2-3} \cmidrule(l{3pt}r{3pt}){4-5} \cmidrule(l{3pt}r{3pt}){6-7} \cmidrule(l{3pt}r{3pt}){8-9}
 & Raw & PH & Raw & PH & Raw & PH & Raw & PH \\
\midrule
Loss  & 1.1153 & 1.1356 & 0.5542 & 0.5629 & 0.8163 & 0.8334 & 0.1986 & 0.2301 \\ 
F1-micro  & 0.3138 & 0.2790 & 0.6229 & 0.6218 & - & - & - & - \\ 
F1-macro  & 0.1873 & 0.1484 & 0.6051 & 0.6008 & - & - & - & - \\  
Precision  & 0.2443 & 0.2172 & 0.6939 & 0.6927 & - & - & - & - \\ 
Recall   & 0.4385 & 0.3900 & 0.5650 & 0.5641 & - & - & - & - \\ 
AUC  & 0.8397 & 0.8226 & 0.8184 & 0.8180 & - & - & - & - \\ 
Acc  & - & - & - & - & 0.7516 & 0.7546 & 0.9454 & 0.9391 \\  
\bottomrule
\end{tabular}
\end{adjustbox}
\caption{Encoder-encoder USE model, all datasets and metrics.}
\label{tbl:enc-enc_use}
\end{table}

\begin{table}[h!]
\begin{adjustbox}{width=\columnwidth,center}
\begin{tabular}[t]{lrrrrrrrr}
\toprule
\multirow{2}{*}{BERTscore} & \multicolumn{2}{c}{CounselChat} & \multicolumn{2}{c}{MH-Chatbot}  & \multicolumn{2}{c}{MentalChat} & \multicolumn{2}{c}{NLP-MH} \\
\cmidrule(l{3pt}r{3pt}){2-3} \cmidrule(l{3pt}r{3pt}){4-5} \cmidrule(l{3pt}r{3pt}){6-7} \cmidrule(l{3pt}r{3pt}){8-9}
 & Raw & PH & Raw & PH & Raw & PH & Raw & PH \\
\midrule
distilgpt2  & 0.8192 & 0.8149 & 0.7646 & 0.7643 & 0.6482 & 0.7899 & 0.7684 & 0.7562 \\
gpt2-medium  & 0.8407 & 0.8404 & 0.7964 & 0.7956 & 0.8023 & 0.8005 & 0.7534 & 0.7650 \\
TinyLlama  & 0.8446 & 0.8482 & 0.8151 & 0.8165 & 0.8100 & 0.8131 & 0.8001 & 0.7940 \\
Qwen   & 0.0840 & 0.8051 & 0.7600 & 0.7523 & 0.7712 & 0.7935 & 0.7844 & 0.7129 \\
\bottomrule
\end{tabular}
\end{adjustbox}
\caption{BERTscore metric for all decoder-decoder models and all datasets.}
\label{tbl:dec-dec_bert}
\end{table}

\begin{table}[h!]
\begin{adjustbox}{width=\columnwidth,center}
\begin{tabular}[t]{lrrrrrrrr}
\toprule
\multirow{2}{*}{ROUGE-L} & \multicolumn{2}{c}{CounselChat} & \multicolumn{2}{c}{MH-Chatbot}  & \multicolumn{2}{c}{MentalChat} & \multicolumn{2}{c}{NLP-MH} \\
\cmidrule(l{3pt}r{3pt}){2-3} \cmidrule(l{3pt}r{3pt}){4-5} \cmidrule(l{3pt}r{3pt}){6-7} \cmidrule(l{3pt}r{3pt}){8-9}
 & Raw & PH & Raw & PH & Raw & PH & Raw & PH \\
\midrule
distilgpt2 & 0.1338 & 0.1292 & 0.0587 & 0.0667 & 0.0659 & 0.0735 & 0.0475 & 0.0355\\
gpt2-medium  & 0.1497 & 0.1465 & 0.1150 & 0.1070 & 0.0996 & 0.0880 & 0.0248 & 0.0336 \\
TinyLlama   & 0.1463 & 0.1450 & 0.1261 & 0.1180 & 0.0844 & 0.0962 & 0.0921 & 0.0747 \\
Qwen  & 0.0127 & 0.0807 & 0.0684 & 0.0718 & 0.0655 & 0.0706 & 0.0799 & 0.0476 \\
\bottomrule
\end{tabular}
\end{adjustbox}
\caption{ROUGE metric for all decoder-decoder models and all datasets.}
\label{tbl:dec-dec_rouge}
\end{table}

\end{document}